\documentclass[10pt,twocolumn,letterpaper]{article}

\usepackage{cvpr}              
\usepackage{caption}
\usepackage{subcaption}
\usepackage{ulem}
\usepackage[dvipsnames]{xcolor}

\definecolor{cvprblue}{rgb}{0.21,0.49,0.74}
\usepackage[pagebackref,breaklinks,colorlinks,citecolor=cvprblue]{hyperref}

\def\paperID{*****} 
\def\confName{CVPR}
\def\confYear{2024}

\title{Benchmarking Monocular 3D Dog Pose Estimation Using In-The-Wild Motion Capture Data}

\author{Moira Shooter\\
{\tt\small m.shooter@surrey.ac.uk}
\and
Charles Malleson\\
{\tt\small charles.malleson@surrey.ac.uk}
\and 
Adrian Hilton\\
{\tt\small a.hilton@surrey.ac.uk}
\and
Centre for Vision, Speech and Signal Processing (CVSSP), University of Surrey,\\
Guildford UK\\
}

\begin{document}
\maketitle
\begin{abstract}
We introduce a new benchmark analysis focusing on 3D canine pose estimation from monocular in-the-wild images. A multi-modal dataset ``3DDogs-Lab'' was captured indoors, featuring various dog breeds trotting on a walkway. It includes data from optical marker-based mocap systems, RGBD cameras, IMUs, and a pressure mat. While providing high-quality motion data, the presence of optical markers and limited background diversity make the captured video less representative of real-world conditions. To address this, we created ``3DDogs-Wild'', a naturalised version of the dataset where the optical markers are in-painted and the subjects are placed in diverse environments, enhancing its utility for training RGB image-based pose detectors. We show that using the 3DDogs-Wild to train the models leads to improved performance when evaluating on in-the-wild data. Additionally, we provide a thorough analysis using various pose estimation models, revealing their respective strengths and weaknesses. We believe that our findings, coupled with the datasets provided, offer valuable insights for advancing 3D animal pose estimation.
\end{abstract}    
\section{Introduction}
\label{sec:intro}
\begin{figure}[tbh]
    \centering
    \includegraphics[width=0.8\linewidth]{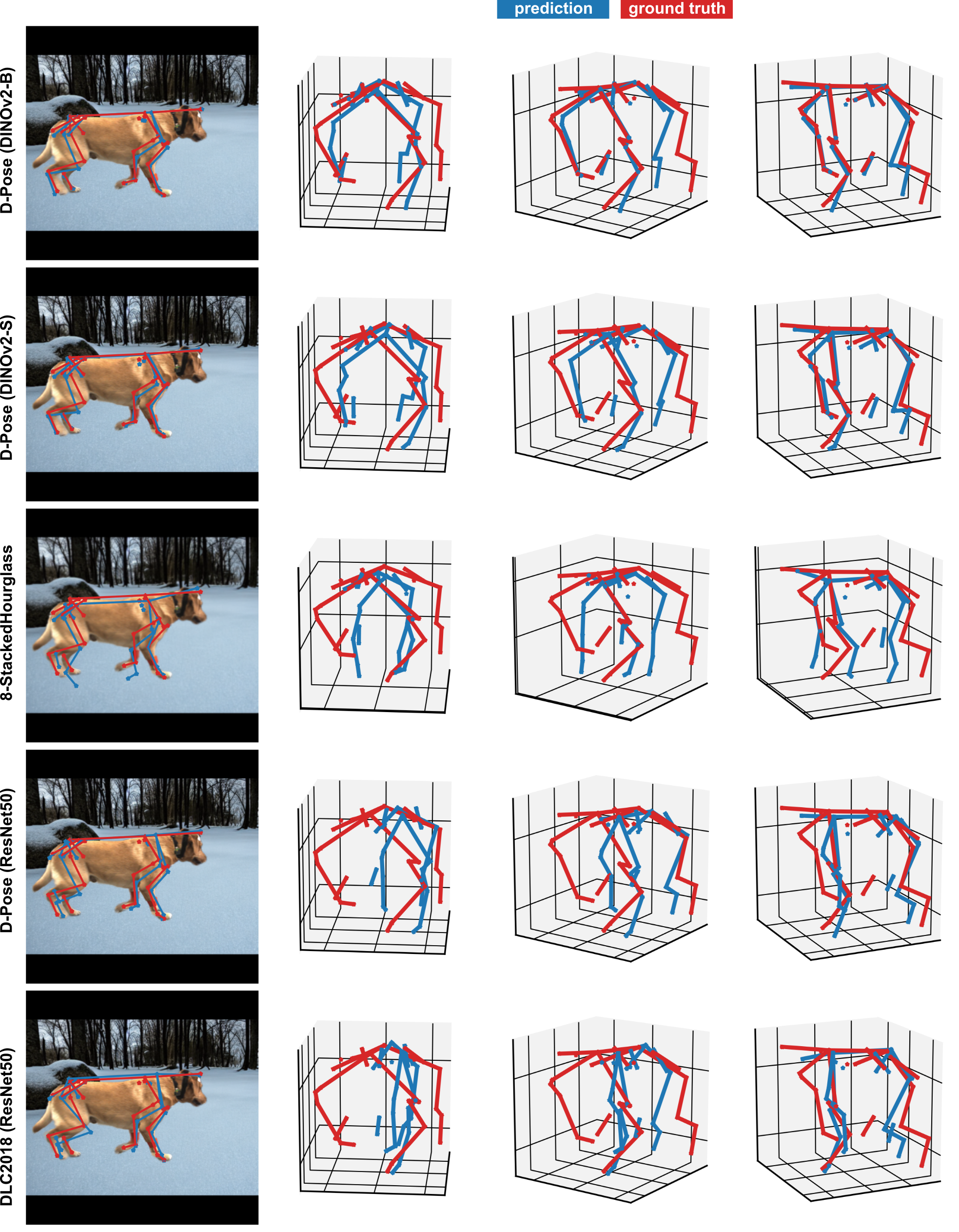}
    \caption{Qualitative results of different pose estimation methods trained on the 3DDogs-Wild dataset.}
    \label{fig:qualitative_results}
\end{figure}
Traditionally, multiple sensor systems such as motion capture (mocap) are used to analyse the gait of animals. While these systems offer accurate kinematic and kinetic information, they are invasive and setting up these systems requires time and specialised expertise. In response to this, there is a growing interest in monocular and markerless approaches using machine learning techniques offering simple and non-invasive systems. While it is possible to use the RGB image data captured by mocap for training pose estimation models, these models are prone to fail to generalise when applied to images captured in natural environments. This is due to the domain gap as there are significant differences in conditions between the controlled environment of motion capture and the varied conditions found in natural settings.

The majority of existing 3D datasets were acquired in controlled environments using multi-view systems \cite{openmonkey,An2023-qv,Deane_2024_WACV,Kearney_2020_CVPR}. 
However, there has been an effort in producing 3D in-the-wild datasets \cite{Xu_2023_ICCV,10.1109/ICRA48506.2021.9561338} where the 3D ground truth was generated by triangulating the 2D joint coordinates from multiple views. Despite this, the reliability of the 3D ground truth is directly dependent on the quality of the 2D manual annotations made by humans.

Recognising the limitations associated with producing 3D datasets, previous research proposed the creation of synthetic datasets using real-time game engines \cite{9157335, sydogVid_Shooter} or employing parametric models \cite{Zuffi:ICCV:2019,Xu_2023_ICCV}. However, a major challenge associated with using synthetic training data is the domain gap between real and synthetic data.
To address this, previous research proposed the generation of near-realistic data using style-transfer methods \cite{spac_net} or by modifying the Grand Theft Auto game to simulate real-world data combined by leveraging existing foundation models \cite{Shooter_2024_WACV}.

Capturing 3D datasets of animals presents greater challenges than working with humans, primarily due to the unpredictability of animal behaviour. Consequently, there's been a growing trend towards releasing 2D animal pose datasets, as they are simpler to develop. These datasets were designed with the aim of both pose estimation and tracking, as well as understanding animal behaviour \cite{yu2021ap,yang2022apt, 9879959,Liu_2023_ICCV}.

Due to these advances and lack of 3D animal pose datasets, previous works related to 3D animal pose relied heavily on 2D ground truth and priors \cite{biggs2020wldo,barc2022rueegg,Zuffi:ICCV:2019,10.3389/fphy.2022.839582}. While these show promising results, due to the sole reliance on 2D signals, there remains a potential for depth ambiguity. Additionally, the acquisition of datasets including prior information such as ground contact \cite{bite2023rueegg} presents a challenge due to the time and cost associated with labelling. Moreover, given the scarcity of 3D datasets, when evaluating the 3D predictions, most of these methods project the 3D pose onto the image plane. This approach does not provide an accurate assessment of the predicted 3D pose.

\textbf{In this paper}, we present a benchmark analysis for monocular 3D dog pose estimation. To leverage the motion capture dataset's 3D ground truth and for machine learning models to generalise across different image conditions, an in-the-wild version of the original dataset was created by removing the optical markers and replacing the default backgrounds. Both datasets will be made available for research.
We focus on the generalisation capabilities of the models using both the mocap and in-the-wild version datasets. Specifically, we explore two key questions: (i) whether the pose estimation models exhibit better performance when trained on the in-the-wild version of the data compared to the original, indoor dataset containing optical markers and no diversity of background. (ii) whether the top performing pose estimation models can generalise to other animal species using the Animal3D dataset \cite{Xu_2023_ICCV}. 
\section{3DDogs-Lab}
\begin{figure}[tbh]
    \centering
    \includegraphics[width=0.8\linewidth]{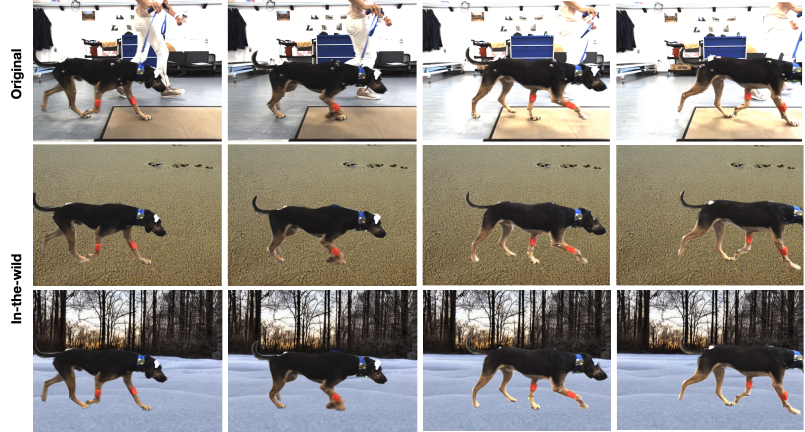}
    \caption{Samples of 3DDogs-Lab showcasing the original motion capture vs in-the-wild version sequences.}
    \label{fig:full_intro}
\end{figure}
\textbf{Data capture of \textit{3DDogs-Lab}}.
The aim of creating the dataset was to detect lameness in dogs by using different types of sensors/capture systems including pressure mat, optical markers, IMUs and RGBD cameras. The capture included 64 dogs, each performing three trials of walking and three of trotting. In light of technical challenges with the capture hardware, it was not possible to obtain valid recordings for all participants. As a result, there is a reduction in the number of usable subjects and trials available for analysis. The final dataset contained a total of 37 subjects and 143 valid recordings. In each trial, the dog moves along a walkway from left to right and then right to left. The capture configuration comprised 8 RGBD cameras along with optical marker and IMU tracking systems. Since all cameras were located on one side of the walkway, only one side of the dog is visible in each pass. Henceforth, the original dataset is denoted as the \underline{\textit{3DDogs-Lab}} dataset.

\textbf{Generation of \textit{3DDogs-Wild}}. To produce an in-the-wild version of the 3DDogs-Lab dataset, we removed the optical markers from the frames and produced various backgrounds for each sequence (\cref{fig:full_intro}). We produced an in-the-wild version for one camera view. As the dog is only in view of each camera for part of the sequence, we only generate the in-the-wild output for frames which the dog is in the camera's field of view. This involved reprojecting the 3D coordinates of optical markers into the image plane. During this step, we also created masks specifically for frames where the optical markers were located within the camera's image plane. These masks were later used in the inpainting process of the optical markers, ensuring a uniform fur appearance without optical markers across frames. By leveraging the capabilities of ProPainter \cite{zhou2023propainter} we achieved near-consistent inpainting results for each recorded sequence. 

To integrate the subjects with various backgrounds, we produced binary masks by combining the binary masks from both off-the-shelf segmentation models such as SAM \cite{kirillov2023segany} and YOLOv8 \cite{Jocher_Ultralytics_YOLO_2023}. The binary masks were merged using operations such as addition and subtraction to leverage the benefits of both. Additionally, we manually verified each frame for quality assurance. The application of binary masks for subject-background composition resulted in a distinct ``copy-paste'' artifact, characterised by overly sharp edges of the subject. We refined the integration of the subject into the background by producing alpha mattes from the binary masks. This process involved creating trimaps, which are segmentation maps delineating foreground, background, and the transitional area, by applying erosion and dilation operations on the binary masks. To produce the final alpha mattes, the RGB image along with the trimap, was fed into the VitMatte image matting model \cite{yao2024vitmatte}.

To generate backgrounds, we leveraged the Stable Diffusion model \cite{yu2023inpaint,Rombach_2022_CVPR}. We originally provided a text prompt, binary mask, and the RGB image as input. However, this method resulted in artifacts, such as additional dog limbs, which were unintended (see Fig. S\ref{fig:background_gen} in supplementary material (Supp.)). While this approach may suffice for a single image, its lack of background consistency rendered it unsuitable for a sequence of frames. Therefore, we decided to generate one static image for each sequence using only a text prompt, such as $t_0=$ \textit{``A clean and empty realistic colored photograph of \{s\} environment with floor.''}. The variable $s$ is set to $s=outdoor$ or $s=indoor$. To enhance the diversity of generated backgrounds, when the variable $s=outdoor$, we expanded the text prompt to include the type of floor, $t=t_0+$\textit{``with floor made of \{e\}''}. The variable $e$ was set to $e=$\{\textit{sand, grass, rocks, pavement, tiles, or snow}\}. A total of 37 subjects, 286 sequences, and 12,940 frames were generated for the final in-the-wild-version of the dataset. Henceforth, the ``naturalised'' version of the dataset is denoted as the \underline{\textit{3DDogs-Wild}} dataset.

\section{Experiments and Results} \label{sec:exp}
\textbf{Evaluation metrics}. We used the Percentage of Correct Keypoints (PCK) and Mean Per Joint Position Error (MPJPE) metrics for evaluating the pose estimation methods. The PCK determines a keypoint to be correctly localised if the predicted position of the keypoint falls within a specified threshold of the ground truth keypoint. Meanwhile, the MPJPE metric calculates the average distance between the predicted and ground truth keypoints. The distances between predicted and ground truth keypoints were normalised relative to the length of the bounding box diagonal for both 2D and 3D PCK. The thresholds for both 2D and 3D PCKs were set to 0.15. While normalisation for 2D MPJPE was based on the bounding box diagonal length, the 3D MPJPE was not normalised and reported in millimeters. Furthermore, we included supplementary 3D metrics, PA-MPJPE and PA-PCK, indicating the metrics after the alignment between predicted 3D pose and 3D ground truth through Procrustes alignment (PA).
\begin{figure}[tbh]
    \centering
    \includegraphics[width=0.8\linewidth]{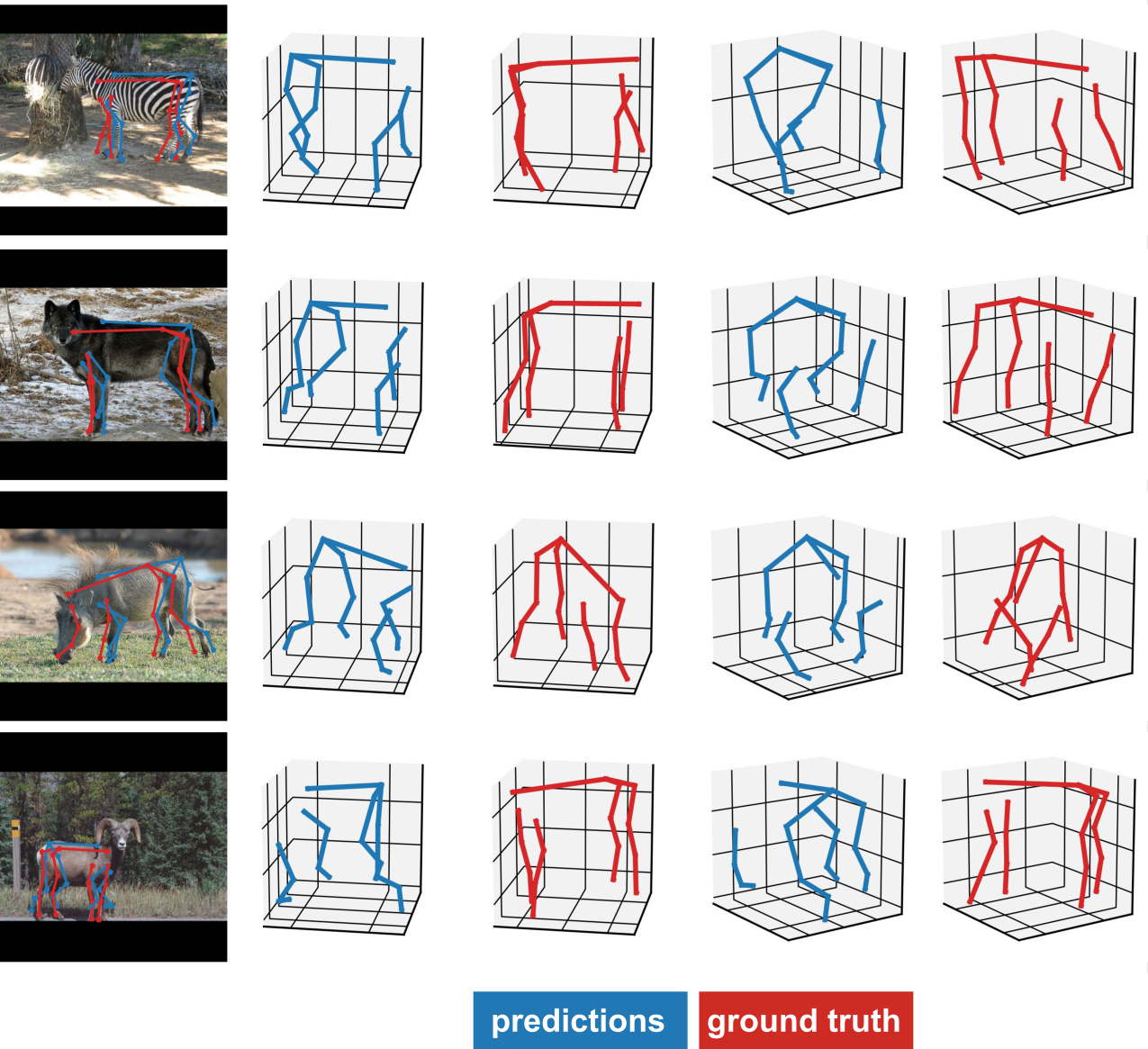}
    \caption{Qualitative results on samples of the Animals3D \cite{Xu_2023_ICCV} test set from D-Pose (DINOv2-S) trained on only the 3DDogs-Wild dataset. The pose is viewed from different angles.}
    \label{fig:Animals3D_benchmark}
\end{figure}
\begin{table*}[]
    \centering 
    \begin{tabular}{@{}lcccc|cc@{}}
    \toprule
     Method & \multicolumn{4}{c}{3D} & \multicolumn{2}{c}{2D} \\
     \midrule
     & MPJPE $\downarrow$ & PA-MPJPE $\downarrow$ & PCK@0.15 $\uparrow$ & PA-PCK@0.15 $\uparrow$ & MPJPE $\downarrow$ & PCK@0.15 $\uparrow$  \\
    \cline{2-7}
    8-StackedHourglass & \underline{20.51} & \underline{14.60} & \uuline{68.04} & \uuline{86.25} & 3.67 & 88.16  \\
    DLC2018 (ResNet50) & 29.79 & 23.79 & 62.35 & 76.65 & \uuline{2.12} & \underline{91.91} \\
    D-Pose (ResNet50)  & \uuline{22.65} & \uuline{16.90} & \underline{75.66} & \underline{91.3}1 & \underline{2.07} & \uuline{90.73}\\
    D-Pose (DINOv2-S) & \textbf{16.81} & \textbf{11.36} & \textbf{86.55} & 98.04 & \textbf{1.56} & \textbf{98.63}\\
    D-Pose (DINOv2-B) & 16.38 & 10.43 & 86.09 & 98.78 & 1.37 & 98.87 \\
    \midrule
    8-StackedHourglass-lab & 39.74 & 24.12 & 32.18 & 61.54 & 10.96 & 45.50 \\
    DLC2018 (ResNet50)-lab & 34.83 & 28.06 & 51.48 & 66.67 & 3.27 & 87.64 \\
    D-Pose (ResNet50)-lab  & 34.78 & 23.03 & 52.53 & 77.09 & 2.98 & 87.33\\
    D-Pose (DINOv2-S)-lab & 19.41 & 11.71 & 80.81 & \textbf{98.12} & 1.72 & 98.39 \\
    D-Pose (DINOv2-B)-lab & 17.83 & 10.90 & 84.65 & 98.63 & 1.62 & 98.55\\
    \bottomrule
    \end{tabular}
    \caption{Quantitative results from different pose estimation networks trained on either the 3DDogs-Wild or 3DDogs-Lab datasets (\textit{-lab}) and \textbf{evaluated on the 3DDogs-Wild} test set. The top-3 best performing networks are highlighted: \textbf{first}, \underline{second}, \uuline{third}, excluding the D-Pose (DINOv2-B) model for fair comparison.}
    \label{tab:poseEstimationbenchmark3D_itwTest}
\end{table*}
\begin{table*}
    \centering
   \begin{tabular}{@{}lcccc|cc@{}}
    \toprule
     Method & \multicolumn{4}{c}{3D} & \multicolumn{2}{c}{2D} \\
     \midrule
     & MPJPE $\downarrow$ & PA-MPJPE $\downarrow$ & PCK@0.15 $\uparrow$ & PA-PCK@0.15 $\uparrow$ & MPJPE $\downarrow$ & PCK@0.15 $\uparrow$ \\
    \cline{2-7}
    8-StackedHourglass & 26.21 & 17.34 & 54.71 & 79.14 & 6.05 & 70.68 \\
    DLC2018 (ResNet50) & 30.65 & 24.74 & 60.62 & 75.72 & 2.73 & 89.50 \\
    D-Pose (ResNet50)  & 24.98 & 18.67 & 72.62 & 88.30 & 2.50 & 89.13\\
    D-Pose (DINOv2-S) & \underline{17.47} & \underline{11.97} & \underline{84.50} & \underline{97.66} & \underline{1.77} & \underline{97.10}\\
    D-Pose (DINOv2-B) & 16.20 & 10.74 & 85.67 & 98.25 & 1.56 & 97.95\\
    \midrule
    8-StackedHourglass-lab & 20.65 & 14.94 & 68.08 & 85.82 & 3.37 & 90.98\\
    D-Pose (ResNet50)-lab  & \uuline{19.15} & \uuline{13.41} & \uuline{82.67} & \uuline{96.00} & \uuline{1.97} & 90.80\\
    DLC2018 (ResNet50)-lab & 24.80 & 19.43 & 73.28 & 86.77 & 2.08 & \uuline{92.14}\\
    D-Pose (DINOv2-S)-lab & \textbf{15.90} & \textbf{10.51} & \textbf{87.28} & \textbf{98.37} & \textbf{1.40} & \textbf{98.83}\\
    D-Pose (DINOv2-B)-lab & 15.87 & 10.57 & 86.43 & 98.64 & 1.30 & 99.08\\
    \bottomrule
    \end{tabular}
    \caption{Quantitative results from different pose estimation networks trained on either the 3DDogs-Wild or 3DDogs-Lab datasets (\textit{-lab}) and \textbf{evaluated on the 3DDogs-Lab} test set. The top-3 best performing networks are highlighted: \textbf{first}, \underline{second}, \uuline{third}, excluding the D-Pose (DINOv2-B) model for fair comparison.}
    \label{tab:poseEstimationbenchmark3D_labTest}
\end{table*}

\textbf{Pose estimation models}. We use 3 different pose estimation networks: an 8-stacked hourglass network \cite{10.1007/978-3-319-46484-8_29}, D-Pose \cite{Shooter_2024_WACV} with varying backbones, and Deeplabcut (DLC2018) \cite{deeplabcut2018}. All networks including a ResNet backbone were fully fine-tuned. However, only the last three layers of the DINOv2 in the D-Pose model underwent fine-tuning. We identified an implementation error in the D-Pose architecture related to the axis permutation. We addressed this issue and document our results using the corrected D-Pose model. Additionally, we modified the resolution of the heatmaps to $64 \times 64$. Originally designed for 2D pose estimation, both the stacked hourglass network and DLC2018 were adapted for 3D pose estimation by extending their methods to output $K\times3$ instead of $K$ heatmaps, where $K=29$ denotes the number of keypoints. The heatmaps represents the probability of the 3D coordinates in $XY$, $ZY$, and $XZ$ planes. The final pose was extracted following the approach outlined in the work of Shooter e al. \cite{Shooter_2024_WACV}. Throughout the experiments, we upheld consistency by applying identical hyperparameters and augmentation strategies (see Sec. S\ref{sup_sec:pose} in Supp.). 


\textbf{Datasets.} The training split, based on breed type, allocated 20 individuals for training, 9 for validation, and 8 for testing, meaning each split contains different individuals. This resulted in 148 sequences for training, 76 for validation, and 62 for testing.

\textbf{Challenges}. (i) \textit{3DDogs-Wild vs 3DDogs-Lab}. We evaluate whether the pose estimation models exhibit better performance when trained on 3DDogs-Wild compared to 3DDogs-Lab. The evaluation of this task is performed on sequences of dogs that were not seen whilst training from the 3DDogs-Wild dataset. Models trained on 3DDogs-Lab are designated with``-lab'' appended to their names. \cref{tab:poseEstimationbenchmark3D_itwTest} shows that models trained with the 3DDogs-Wild dataset demonstrate enhanced performance compared to those trained with 3DDogs-Lab. One might argue this is due to the domain gap. However, when evaluating these models on cross-domain test sets, we observe a narrowing of the performance gap. Models trained on 3DDogs-Wild exhibit more robust generalisation to the 3DDogs-Lab dataset than vice versa (\cref{tab:poseEstimationbenchmark3D_itwTest} vs \cref{tab:poseEstimationbenchmark3D_labTest}). For example, when comparing the MPJPE performance of both the 8-StackedHourglass networks evaluated on the 3DDogs-Wild test set in \cref{tab:poseEstimationbenchmark3D_itwTest}, the difference in performance is equal to 19.23\%. However, when comparing the same models and metric evaluated on 3DDogs-Lab test set in \cref{tab:poseEstimationbenchmark3D_labTest} the difference in performance is equal to 5.56\%. This confirms our initial hypothesis that models exhibit superior generalisation capabilities to in-the-wild data and demonstrate enhanced performance on out-of-domain datasets when trained on 3DDogs-Wild data compared to mocap data (3DDogs-Lab).
\begin{table*}
\small
    \centering
    \begin{tabular}{@{}lcccc|cc@{}}
    \toprule
     Method & \multicolumn{4}{c}{3D} & \multicolumn{2}{c}{2D} \\
     \midrule
     & MPJPE $\downarrow$ & PA-MPJPE $\downarrow$ & PCK@0.15 $\uparrow$ & PA-PCK@0.15 $\uparrow$ & MPJPE $\downarrow$ & PCK@0.15 $\uparrow$  \\
    \cline{2-7}
    8-StackedHourglass-anim & \uuline{153.43}&  129.14& 28.17& 27.29& 20.41&27.12 \\
    D-Pose (ResNet50)-anim & 160.60 & 123.84 &\underline{30.24}&29.88&\underline{10.31}&61.00\\
    D-Pose (DINOv2-S)-anim &  169.85 & \textbf{108.79} & 27.16 & \textbf{37.31} & \textbf{5.98} & \textbf{88.33}\\
    \midrule
    8-StackedHourglass* & 159.03&128.90&24.17&27.72&26.46&24.46\\
    D-Pose (ResNet50)* & 159.74& 129.71&25.09&25.98&15.04& 46.15\\
    D-Pose (DINOv2-S)* & \textbf{149.54} & \uuline{119.52}&\textbf{30.63}&\underline{33.15}&\uuline{10.46}&\underline{68.39} \\
    \midrule
    8-StackedHourglass-lab* & 163.36&128.47&19.84&27.21&30.87&12.58\\
    D-Pose (ResNet50)-lab* &161.52&129.76&23.40&26.52&17.96&36.69\\
    D-Pose (DINOv2-S)-lab* &\underline{151.14}&\underline{118.86}&\uuline{29.80}&\uuline{33.13}&11.32&\uuline{64.13} \\
    \bottomrule
    \end{tabular}
    \caption{Quantitative results \textbf{evaluated on the Animals3}D \cite{Xu_2023_ICCV} test set using top performing models trained on Animals3D (\textit{-anim}), 3DDogs-Wild and 3DDogs-Lab datasets (\textit{-lab}) for both 3D and 2D. D-Pose (DINOv2-B) is excluded for fair comparison. The top-3 best performing networks are highlighted: \textbf{first}, \underline{second}, \uuline{third}. The asterisk (*) denotes that the model was trained on a subset of the 3DDogs dataset, proportionate to the Animals3D dataset.}
    \label{tab:benchmark3DAnimals}
\end{table*}

To compare the models fairly, the D-Pose (DINOv2-B) model was excluded due to its high parameter count (see Tab. S\ref{tab:number_of_par} in Supp.). After assessing the models with the 3DDogs-Lab test set, D-Pose (DINOv2-S) with lower parameter count emerged as the top performer. In terms of 3D pose evaluation, both the 8-StackedHourglass and D-Pose (ResNet50) models are competing for the second position, while in the 2D pose assessment, the DLC2018 (ResNet50) and D-Pose (ResNet50) models are competing for the next-best position. Overall, we argue that D-Pose (ResNet50) is the second best performing model due to its consistent performance across both 3D and 2D evaluations, as well as its strong performance on the cross-domain test set, where the domain gap is narrower. When comparing the ``raw'' and PA-metrics, D-Pose (DINOv2) models demonstrate strong performance even without requiring alignment. These results, underlines the significance of selecting and leveraging effective pre-trained backbones to attain high performance in animal pose estimation tasks. Furthermore, when comparing the D-Pose models with varying sizes of DINOv2 backbones, as expected, there is an improvement in performance as the number of parameters increases. 

(ii) \textit{Generalisation to other species}. We evaluate whether the top-3 best performing pose estimation models from the previous challenge can generalise to other animal species using the Animal3D \cite{Xu_2023_ICCV} dataset. We first perform a baseline with the models trained on the Animals3D dataset and then evaluate the models that were trained either on the 3DDogs-Wild or the 3DDogs-Lab dataset. To train the models with the Animals3D dataset, the dataset was split with an 80:20 training/testing ratio. Given that the Animals3D dataset comprises $\sim$3K data points, which is smaller in comparison to the $\sim$12K data points in the 3DDogs dataset, we present results from models trained on a subset of the 3DDogs dataset equal in size to the Animals3D dataset to have a fair comparison.

The D-Pose (DINOv2-S)-anim model trained on the Animals3D dataset demonstrates superior performance in 2D, outperforming other models trained on the same dataset. While other models trained on Animals3D exhibit stronger performance in raw metrics for 3D, D-Pose (DINOv2-S)--anim outperforms them after applying Procrustes alignment. In terms of overall performance, D-Pose (DINOv2-S) trained on the 3DDogs-Wild dataset stands out as the best model, particularly when assessing raw metrics against all other models. The reason why D-Pose (DINOv2-S) trained on the 3DDogs-Wild dataset does not achieve comparable 2D performance to D-Pose (DINOv2-S)-anim trained on the Animals3D dataset is due to the variation in viewpoints present in the Animals3D dataset, whereas the 3DDogs dataset features samples with consistent viewpoints. Additionally, it can be concluded that when looking at the raw metrics (prior to Procrustes alignment), all the models trained on the 3DDogs-Wild dataset generalise better to out-of-domain data than models trained on mocap data (3DDogs-lab) for both 3D and 2D. 

The following section discusses the models trained on the 3DDogs-Wild dataset. Both the 8-StackedHourglass and D-Pose (ResNet50) models fail to generalise to cross-domain data. The D-Pose model exhibits superior performance due to the generalisation capabilities of DINOv2, however, its performance on the Animals3D test set falls short for 3D and remains acceptable for 2D (\cref{tab:benchmark3DAnimals}). However, qualitative evaluation shows that D-Pose predicts plausible 3D poses (\cref{fig:Animals3D_benchmark}). This is due to 3 reasons: 
\begin{enumerate}
    \item The differing keypoint semantics between the 3DDogs-Lab and Animals3D datasets (see Fig. S\ref{fig:keypoint_diff} in Supp.)
    \item The Animals3D dataset features images of animals facing the camera which poses a challenge, as the model was exclusively trained on side-view dog images 
    \item The Animals3D dataset was produced by fitting the animal parametric model, SMAL \cite{Zuffi_2017_CVPR}, to monocular images using only 2D signals. Despite this process, there remains a potential for depth ambiguity, indicating that the resulting 3D representations may not fully capture the true ground truth
\end{enumerate}


\section{Conclusion}
We present a new benchmark and two datasets for 3D dog pose estimation from monocular views, designed to facilitate fair comparisons of pose estimation methods across various datasets. 3DDogs-Wild offers both in-the-wild and 3D gold standard ground truth, sourced from the 3DDogs-Lab dataset. We offer valuable insights into the capabilities and limitations of existing pose estimation techniques, by addressing key questions regarding model performance with in-the-wild data compared to motion capture data, as well as generalisation across different animal species. Our results highlight the superiority of D-Pose over alternative models, attributed to its backbone and we show that models trained with 3DDogs-Wild have better generalisation and performance compared to those trained on 3DDogs-Lab due to the increased diversity in the dataset. 
\subsection*{Acknowledgement}
This research was partially supported by EPSRC Audio-Visual Media Platform Grant EP/P022529/1 and by the Leverhulme Trust Early Career Fellowship scheme.
{
    \small
    \bibliographystyle{ieeenat_fullname}
    \bibliography{main}
}

\clearpage
\setcounter{page}{1}
\maketitlesupplementary

\section{3DDogs-Lab}
\subsection{Generation of 3DDogs-Wild}
\label{sec:supp_gen}
\begin{figure}[tbh]
    \centering
    \includegraphics[width=\linewidth]{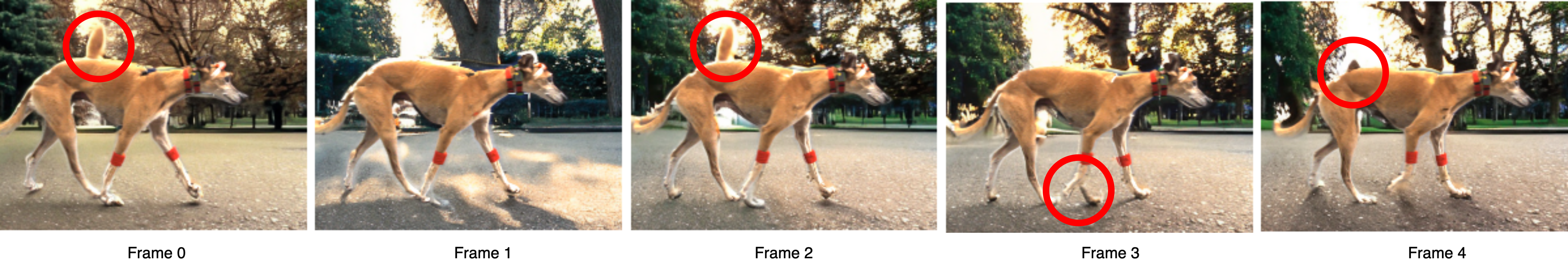}
    \caption{Background generation artifacts such as extra limb hallucinations and inconsistencies across frames.}
    \label{fig:background_gen}
\end{figure}

\section{Experiments and Results}
\subsection{Pose estimation models} \label{sup_sec:pose}
The models were optimised using an Adam optimizer with a batch size of 8. Initially, the learning rate was established at 3e-5, which was subsequently reduced by a factor of 1/30 after 7 epochs. Although the maximum number of epochs was set to 300, we implemented early stopping for enhanced training effectiveness, halting when the validation loss ceased to decrease for 5 consecutive epochs. During training, data augmentation was performed by randomly applying color jitter, gaussian blur, and grayscale transformations with a probability of 0.3 for each operation.

\subsection{3DDogs-Wild vs 3DDogs-Lab}
In \cref{sec:exp} we compared various pose estimation models, however, often we exclude the D-Pose (DINOv2-B) model due to high parameter count, as we expected it to yield superior performance compared to models with fewer parameters. \cref{tab:number_of_par} shows the pose estimation models including the number of parameters. 
\begin{table}[bht]
    \centering
    \begin{tabular}{@{}lc@{}}
    \toprule
     Method & parameters (M) \\
     \midrule
     8-StackedHourglass & 25.9 \\
     DLC2018 (ResNet50) & 26.7 \\ 
     D-Pose (ResNet50) & 46.7 \\
     D-Pose (DINOv2-S) & 22.9 \\
     D-Pose (DINOv2-B) & 88.2 \\
    \bottomrule
    \end{tabular}
    \caption{Number of parameters for each pose estimation model.}
    \label{tab:number_of_par}
\end{table}

Additionally, we present the mean inference time using 170 samples (\cref{tab:infernecetime}). The inference time was measured on hardware with a single GPU, an NVIDIA GeForce RTX 3060 with 8GB of memory. While the D-Pose (DINOv2-S) model is marginally slower than the D-Pose (ResNet50), its significantly improved performance justifies this trade-off. For a well-balanced compromise between accuracy and speed, the D-Pose (ResNet50) or D-Pose (DINOv2-S) models should be considered. However, if utmost accuracy is the priority, one should be mindful of the longer inference time associated with D-Pose (DINOv2-B).
\begin{table}[bht]
    \centering
    \begin{tabular}{@{}lc@{}}
    \toprule
     Method & Mean duration (ms) \\
     \midrule
     8-StackedHourglass & 66.05  \\
     DLC2018 (ResNet50) & 17.39 \\ 
     D-Pose (ResNet50) & 24.28 \\
     D-Pose (DINOv2-S) & 27.91 \\
     D-Pose (DINOv2-B) & 59.92 \\
    \bottomrule
    \end{tabular}
    \caption{Mean inference time for each model using 170 samples.}
    \label{tab:infernecetime}
\end{table}

\begin{figure*}
    \centering
    \includegraphics[width=\linewidth]{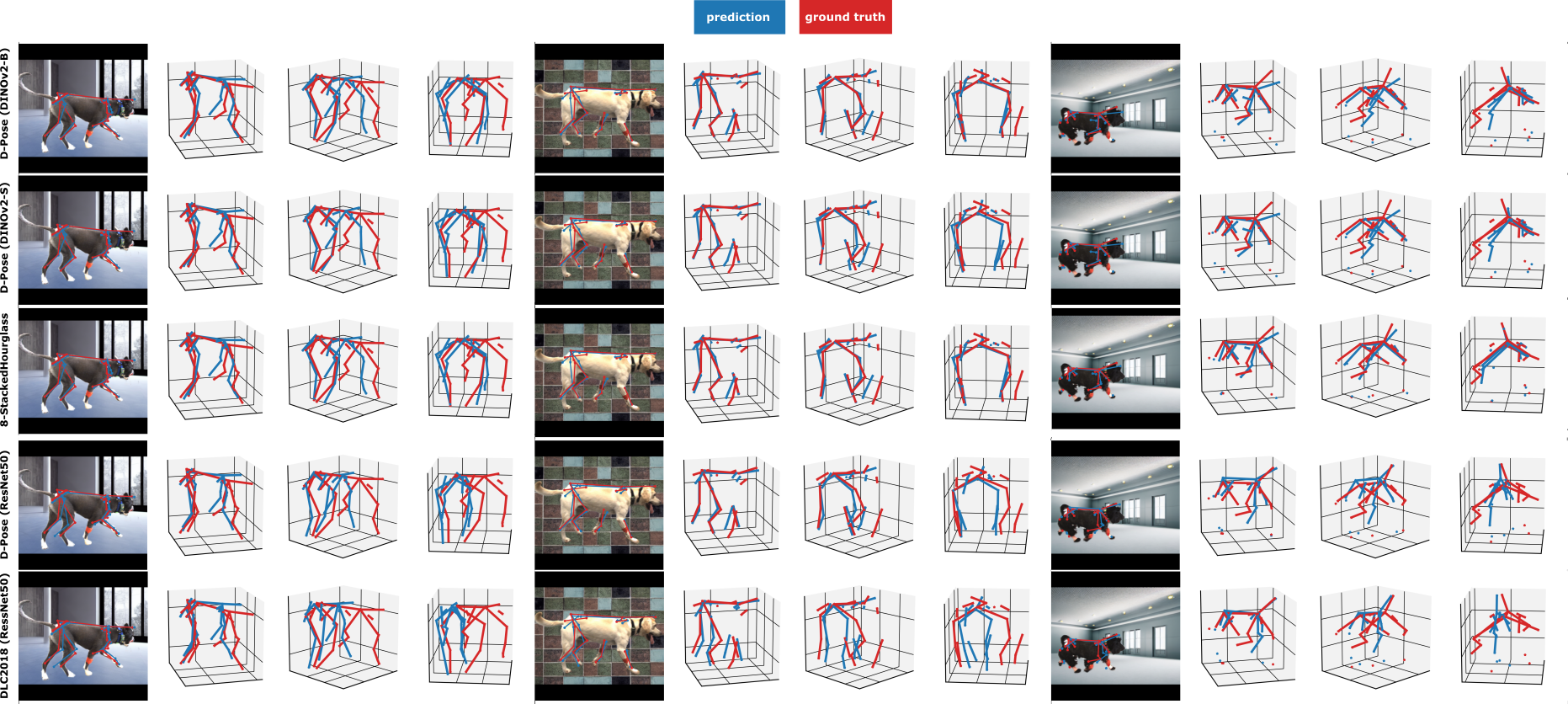}
    \caption{Further qualitative results on samples of the 3DDogs-Wild test set from different pose estimation models.}
    \label{fig:qualitative_further_itw}
\end{figure*}

We show additional qualitative results in \cref{fig:qualitative_further_itw} from samples of the 3DDogs-Wild test set. Technical challenges during capturing the dataset, such as subjects shaking off the optical markers, lead to incomplete 3D poses in our results. 

\subsection{Generalisation to other species}
\begin{figure}
    \centering
    \includegraphics[width=\linewidth]{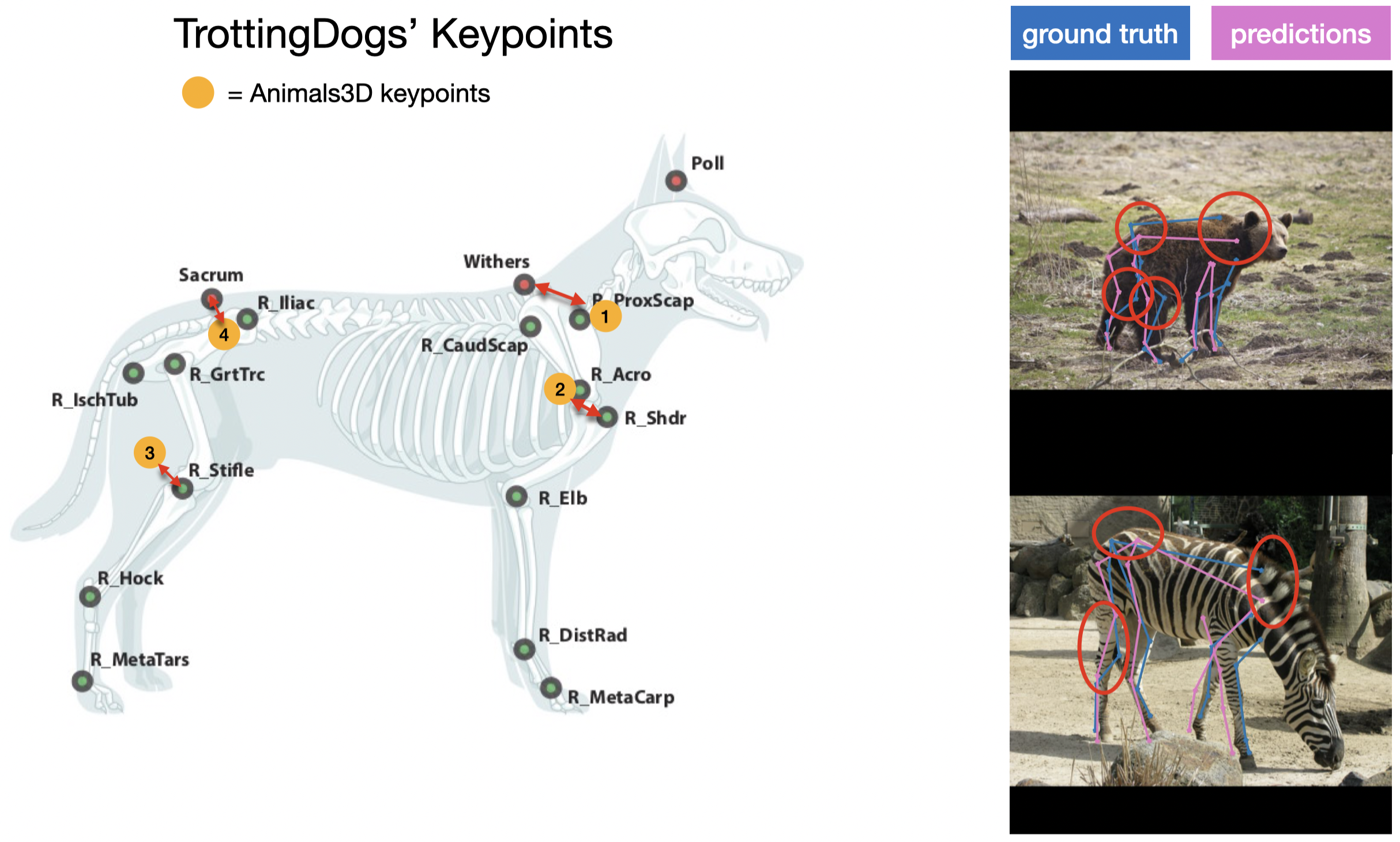}
    \caption{Demonstrating the keypoint differences between the 3DDogs and Animals3D datasets. Additionally, showcasing samples of D-Pose (DINOv2-S) model's predictions and ground truth of the Animals3D test set. }
    \label{fig:keypoint_diff}
\end{figure}

In \cref{fig:keypoint_diff} we show the definitions of the keypoints in the 3DDogs-Lab dataset overlayed with the Animals3D's keypoint definitions that are semantically different from the 3DDogs-Lab dataset. 
To demonstrate the differences between keypoint semantics we visualise it using red arrows. 

\begin{figure*}
    \centering
    \includegraphics[width=\linewidth]{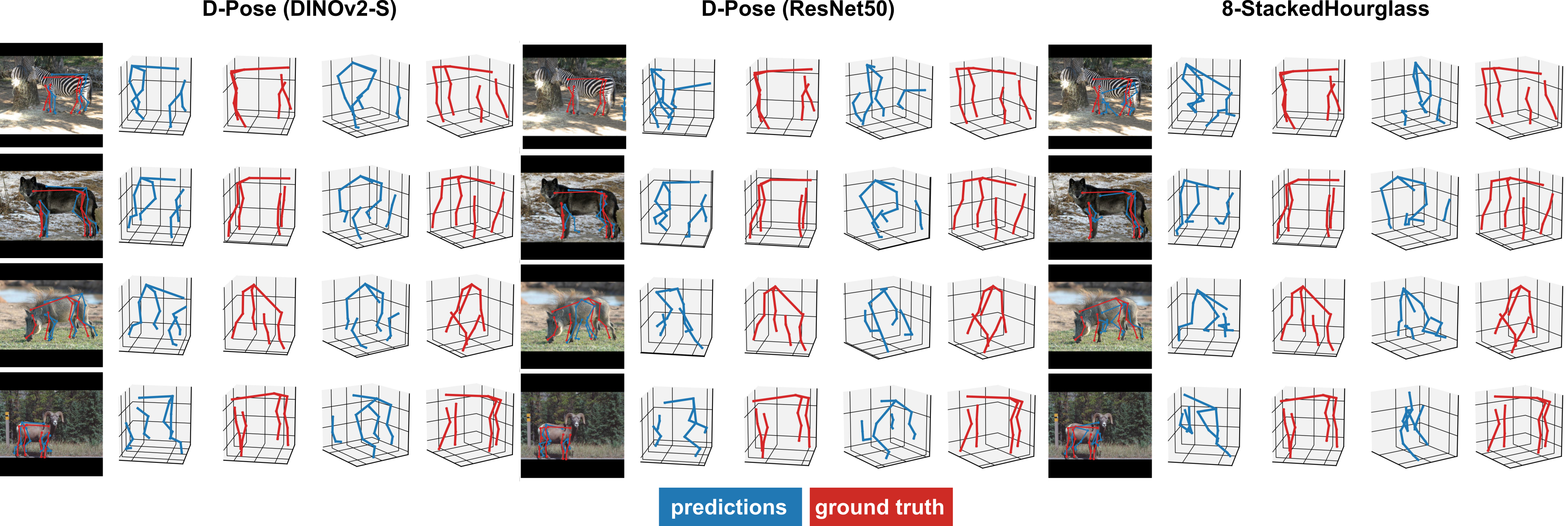}
    \caption{Qualitative results evaluated on samples of the Animals3D dataset \cite{Xu_2023_ICCV} from the top-performing networks trained solely on the 3DDogs-Wild dataset.}
    \label{fig:qualitative_compare}
\end{figure*}

\begin{figure*}
    \centering
    \includegraphics[width=\linewidth]{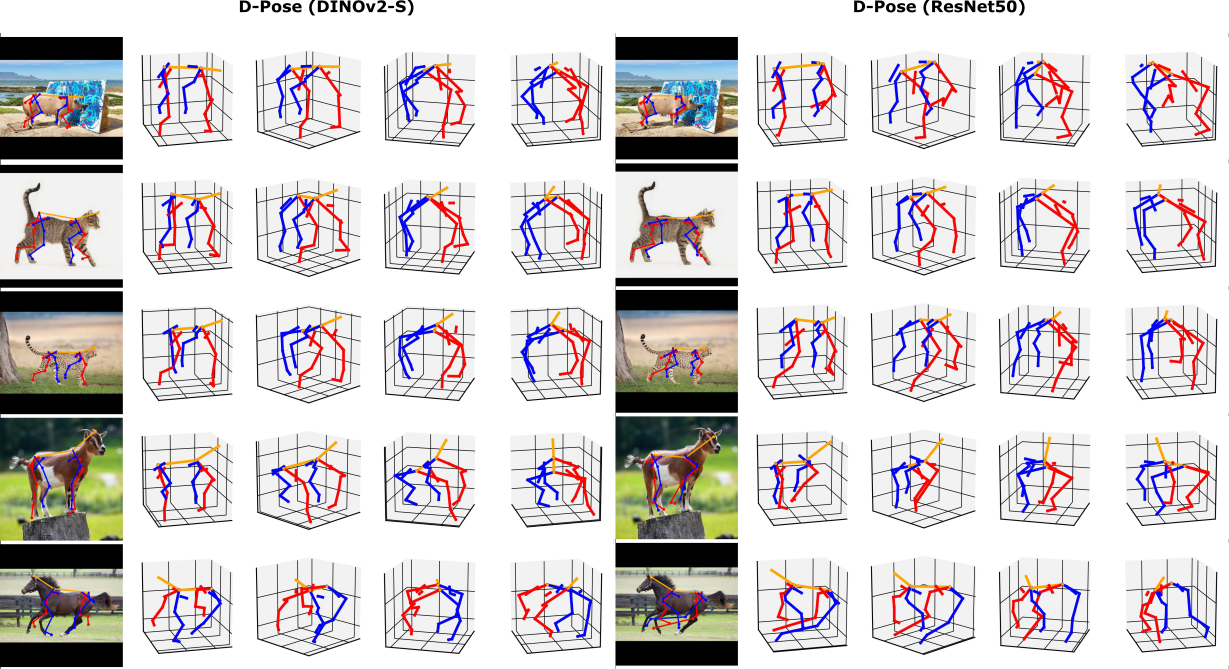}
    \caption{Qualitative results on random internet images from the 2-top best performing models.}
    \label{fig:animals3D_qual}
\end{figure*}

In \cref{fig:qualitative_compare} we show qualitative results for samples of the Animals3D test set comparing the top-3 performing networks e.g. D-Pose (DINOv2-S), D-Pose (ResNet50) and 8-StackedHourglass. We provide further qualitative results from the 2-top performing networks in \cref{fig:animals3D_qual}, showcasing random images or videos sourced from the internet. For the videos, please refer to the supplementary material folder. We estimate the pose on a frame-basis. We omitted results from the stacked hourglass network, as indicated in \cref{tab:benchmark3DAnimals}, due to its inability to generalise to cross-domain data. In the visualisations, the right side of the skeleton is coloured in red, while the left side is coloured in blue. Upon global evaluation both D-Pose (DINOv2-S) and D-Pose (ResNet50) demonstrate the ability to generate plausible 3D poses across various animal species.  However, a clear distinction emerges when the subject appearance differs significantly from that of a dog, as seen with the samples of horses. In such cases, while D-Pose (DINOv2-S) maintains accuracy, D-Pose (ResNet50) fails to yield convincing results. This discrepancy is further highlighted in the supplementary videos accompanying the paper.

\end{document}